\documentclass{article}
\usepackage{spconf,amsmath,graphicx,hyperref}
\usepackage{booktabs}
\usepackage{multirow}
\usepackage{xcolor}

\title{Training-Free and Interpretable Hateful Video Detection via Multi-stage Adversarial Reasoning}
\name{%
  Shuonan~Yang$^{1}$,
  Yuchen~Zhang$^{2}$,
  Zeyu~Fu$^{1*}$\thanks{*Corresponding author. This work was supported by the Alan Turing Institute and DSO National Laboratories Framework Grant Funding.}
}

\address{%
  $^{1}$Multimodal Intelligence Lab, Department of Computer Science, University of Exeter, United Kingdom\\
  $^{2}$Institute for Analytics and Data Science, University of Essex, United Kingdom
}

\begin{document}
\ninept
\maketitle
%
\begin{abstract}
Hateful videos pose serious risks by amplifying discrimination, inciting violence, and undermining online safety. 
Existing training-based hateful video detection methods are constrained by limited training data and lack of interpretability, while directly prompting large vision-language models often struggle to deliver reliable hate detection. To address these challenges, this paper introduces \textit{MARS}, a training-free \textbf{M}ulti-stage \textbf{A}dversarial \textbf{R}ea\textbf{S}oning framework that enables reliable and interpretable hateful content detection \footnote{https://github.com/Multimodal-Intelligence-Lab-MIL/MARS}. \textit{MARS} begins with the objective description of video content, establishing a neutral foundation for subsequent analysis. Building on this, it develops evidence-based reasoning that supports potential hateful interpretations, while in parallel incorporating counter-evidence reasoning to capture plausible non-hateful perspectives. Finally, these perspectives are synthesized into a conclusive and explainable decision. Extensive evaluation on two real-world datasets shows that \textit{MARS} achieves up to 10\% improvement under certain backbones and settings compared to other training-free approaches and outperforms state-of-the-art training-based methods on one dataset. In addition, \textit{MARS} produces human-understandable justifications, thereby supporting compliance oversight and enhancing the transparency of content moderation workflows. 
\end{abstract}
{\textcolor{red}{\textbf{Warning}: This paper contains sensitive materials that many will find hateful or offensive; however, this is necessary due to the nature of the task.\\}}

\section{Introduction}
\label{sec:intro}
Online video-sharing platforms have become a primary medium for information dissemination, with its unprecedented reach enabling the rapid propagation of hateful content \cite{ yang2025temporalnoise, wu2025confront}. YouTube reports approximately 2.6 million new video uploads daily, of which 192,856 were removed for hate speech in the first quarter of 2025 alone \cite{youtube_videos_2025, google_tr_youtube_removals_2025}. At this scale, multimodal hateful video detection has become a high-risk, compliance-critical issue where accuracy, robustness, and transparency are indispensable requirements. 

Hateful content in videos manifests through complex combinations of spoken language, visual imagery, on-screen text, vocal intonation, and contextual symbols, often requiring sophisticated cross-modal reasoning for identification \cite{koushik2025towards, mandal2024attentivefusiontransformerbasedapproach, mariconti2019you}. Das et al. introduced the first hateful video dataset, HateMM \cite{das2023hatemm}, and established a standard pipeline for representation-level fusion classification using features from video frames, audio, and speech transcriptions. Subsequently, Wang et al. constructed the MultiHateClip (MHC) \cite{wang2024multihateclip} dataset, encompassing Chinese and English videos to address the gap in multilingual hateful video datasets. 
Later studies advanced classification performance through cross-modal attention architectures, such as CMFusion \cite{CMFusion}, and retrieval-augmented models, such as MoRE \cite{MoRE}, which enhance contextual understanding by retrieving semantically similar samples from memory banks.

Whilst these training-based approaches optimise learning for known patterns, they are constrained by data scarcity \cite{mathew2021hatexplain}. Unlike purely textual hate speech corpora containing millions of annotated examples, video datasets are limited in scale with prohibitive expansion costs. HateMM contains merely 1,083 videos, whilst MHC comprises approximately 2,000, falling substantially short of the scale required for training generalizable deep models. 
Lack of interpretability further constrains deployment. Most multimodal detectors function as black-box systems outputting binary labels, providing at most attention visualisations or feature importance analyses. Current fusion-centric architectures \cite{CMFusion,das2023hatemm, MoRE}  are not inherently designed to offer transparent and explainable predictions. 
With the EU AI Act \cite{eu_ai_act_2024} and related regulations elevating interpretability and accountability requirements for high-risk applications, it is crucial for content moderation systems to present human-readable decision rationales, uncertainty quantification, and specific cross-modal cues driving decisions. 
\begin{figure*}[t]
    \centering
    \includegraphics[width=0.95\textwidth]{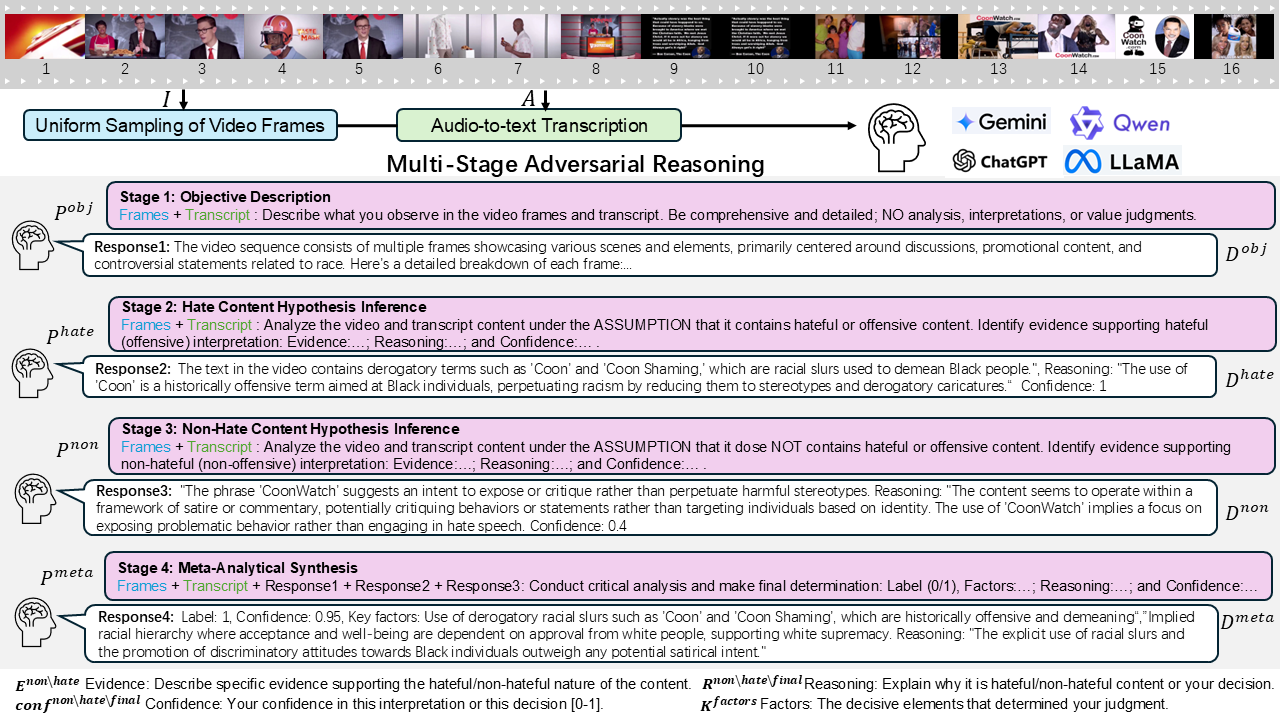}
    \caption{Overview of the proposed multi-stage adversarial reasoning framework. 
}
    \label{fig:framework}
\end{figure*}


Recent work by Wang et al. \cite{Wang2025} conducted preliminary explorations of hateful video detection with VLMs, yet even state-of-the-art models exhibited low accuracy and weak precision, largely due to high false-positive rates. Thus, despite the importance of this task, no study has systematically addressed the reduction of false positives in training-free VLMs for hateful video detection. Meanwhile, Chain-of-Thought (CoT) prompting \cite{CoT2023, cot2024} has been proposed to enhance interpretability, calibration, and task performance by introducing intermediate reasoning steps. InstructMemeCL \cite{InstructMemeCL}, for example, employs CoT to improve task accuracy and interpretability. However, growing evidence shows that CoT explanations often constitute post hoc rationalisations lacking causal grounding \cite{CoTnotExplainability, cotnotexplain2, cotnotexplain3}, and therefore fail to provide genuine auditability.

To address these limitations, we propose a training-free \textbf{M}ulti-stage \textbf{A}dversarial \textbf{R}ea\textbf{S}oning (\textit{MARS}) framework based on VLMs, fundamentally requiring the model to provide the strongest evidence for both hateful and non-hateful hypothesis spaces before making comprehensive judgement, as shown in Figure \ref{fig:framework}. 
\textit{MARS} leverages the reasoning capabilities of VLMs while addressing the limitations of CoT approaches, grounding its decisions in explicit evidence comparison and thereby delivering robust and interpretable hateful video detection.

The main contributions of our paper are threefold:
    (1) To the best of our knowledge, \textit{MARS} is the first interpretable, training-free hateful video detection framework that leverages the intrinsic reasoning capabilities of VLMs, thereby overcoming the limitations of training-based approaches constrained by scarce critical data.
     (2) \textit{MARS} delivers detection results accompanied by human-understandable, auditable justifications, supported by explicit evidence chains that trace how multimodal cues are weighed and synthesized, thereby enhancing both transparency and compliance readiness.
    (3) Extensive evaluations on two real-world datasets demonstrate that \textit{MARS} consistently outperforms existing training-free methods in both accuracy and precision metrics with maximum improvements exceeding 10\% under certain backbones and settings, while maintaining competitive performance against training-based models.

\begin{table*}[ht]
\setlength{\tabcolsep}{4pt} 
\caption{Video classification performance (\%). Acc: Accuracy; MF1: Macro-F1; F1: F1 for hate class; P: Precision for hate class; R: Recall for hate class. English Dataset: HateMM \cite{das2023hatemm}; Chinese Dataset: MultiHateClip \cite{wang2024multihateclip}. \textbf{Bold} indicates best performance, \underline{underlined} indicates second-best performance for each metric, with comparisons among training-free methods.}
\label{tab:results}
\centering
\small
\begin{tabular}{lccccc|ccccc}
\toprule
\multirow{2}{*}{\textbf{Model}} 
& \multicolumn{5}{c|}{\textbf{English Dataset}} 
& \multicolumn{5}{c}{\textbf{Chinese Dataset}} \\
& Acc & MF1 & F1 & P & R 
& Acc & MF1 & F1 & P & R \\
\midrule

\multicolumn{11}{l}{\textit{Training Based}} \\
Basline 
& 79.8\textsubscript{\scriptsize $\pm$0.023} & 78.9\textsubscript{\scriptsize $\pm$0.020} & 75.9\textsubscript{\scriptsize $\pm$0.021} & 73.6\textsubscript{\scriptsize $\pm$0.715} & 78.7\textsubscript{\scriptsize $\pm$0.886} 
& 68.3\textsubscript{\scriptsize $\pm$0.023} & 64.0\textsubscript{\scriptsize $\pm$0.027} & 51.8\textsubscript{\scriptsize $\pm$0.062} & 51.8\textsubscript{\scriptsize $\pm$0.033} & 53.1\textsubscript{\scriptsize $\pm$0.116} \\
CMFusion 
& 80.2\textsubscript{\scriptsize $\pm$0.020} & 79.1\textsubscript{\scriptsize $\pm$0.018} & 74.4\textsubscript{\scriptsize $\pm$0.017} & 77.1\textsubscript{\scriptsize $\pm$0.040} & 72.2\textsubscript{\scriptsize $\pm$0.050} 
& 68.6\textsubscript{\scriptsize $\pm$0.021} & 61.4\textsubscript{\scriptsize $\pm$0.022} & 44.9\textsubscript{\scriptsize $\pm$0.040} & 53.5\textsubscript{\scriptsize $\pm$0.061} & 39.3\textsubscript{\scriptsize $\pm$0.052} \\
MoRE 
& 82.1\textsubscript{\scriptsize $\pm$0.022} & 81.0\textsubscript{\scriptsize $\pm$0.020} & 76.6\textsubscript{\scriptsize $\pm$0.021} & 80.0\textsubscript{\scriptsize $\pm$0.056} & 73.5\textsubscript{\scriptsize $\pm$0.062} 
& 69.6\textsubscript{\scriptsize $\pm$0.021} & 60.2\textsubscript{\scriptsize $\pm$0.023} & 40.8\textsubscript{\scriptsize $\pm$0.035} & 57.1\textsubscript{\scriptsize $\pm$0.037} & 31.7\textsubscript{\scriptsize $\pm$0.055} \\

\midrule
\multicolumn{11}{l}{\textit{Training Free — Simple Prompt}} \\
Qwen2.5  
& 70.3\textsubscript{\scriptsize $\pm$0.018} & 70.2\textsubscript{\scriptsize $\pm$0.018} & 70.9\textsubscript{\scriptsize $\pm$0.021} & 58.2\textsubscript{\scriptsize $\pm$0.029} & 90.8\textsubscript{\scriptsize $\pm$0.029} 
& 70.3\textsubscript{\scriptsize $\pm$0.036} & 60.8\textsubscript{\scriptsize $\pm$0.029} & 41.6\textsubscript{\scriptsize $\pm$0.031} & 59.8\textsubscript{\scriptsize $\pm$0.075} & 32.4\textsubscript{\scriptsize $\pm$0.045} \\
LLaMA4      
& 69.8\textsubscript{\scriptsize $\pm$0.032} & 69.8\textsubscript{\scriptsize $\pm$0.033} & 70.7\textsubscript{\scriptsize $\pm$0.031} & 57.8\textsubscript{\scriptsize $\pm$0.041} & 91.2\textsubscript{\scriptsize $\pm$0.016} 
& 68.7\textsubscript{\scriptsize $\pm$0.020} & 58.9\textsubscript{\scriptsize $\pm$0.028} & 38.7\textsubscript{\scriptsize $\pm$0.049} & 63.3\textsubscript{\scriptsize $\pm$0.045} & 27.9\textsubscript{\scriptsize $\pm$0.069} \\
GPT5 & 
68.4\textsubscript{\scriptsize $\pm$0.032} & 
68.1\textsubscript{\scriptsize $\pm$0.032} & 
71.1\textsubscript{\scriptsize $\pm$0.031} & 
55.9\textsubscript{\scriptsize $\pm$0.035} & 
97.9\textsubscript{\scriptsize $\pm$0.014} & 
68.7\textsubscript{\scriptsize $\pm$0.036} & 
63.3\textsubscript{\scriptsize $\pm$0.036} & 
49.1\textsubscript{\scriptsize $\pm$0.043} & 
54.7\textsubscript{\scriptsize $\pm$0.042} & 
44.6\textsubscript{\scriptsize $\pm$0.044} \\
Gemini2.5 & 
62.7\textsubscript{\scriptsize $\pm$0.027} & 
61.6\textsubscript{\scriptsize $\pm$0.029} & 
67.7\textsubscript{\scriptsize $\pm$0.027} & 
51.6\textsubscript{\scriptsize $\pm$0.029} & 
\underline{98.5\textsubscript{\scriptsize $\pm$0.014}} & 
67.3\textsubscript{\scriptsize $\pm$0.037} & 
64.9\textsubscript{\scriptsize $\pm$0.037} & 
55.7\textsubscript{\scriptsize $\pm$0.042} & 
51.3\textsubscript{\scriptsize $\pm$0.029} & 
60.9\textsubscript{\scriptsize $\pm$0.064} \\
\midrule
\multicolumn{11}{l}{\textit{Training Free — Chain-of-Thought Prompt}} \\
Qwen2.5
& 72.1\textsubscript{\scriptsize $\pm$0.008} & 72.0\textsubscript{\scriptsize $\pm$0.009} & 73.5\textsubscript{\scriptsize $\pm$0.008} & 59.0\textsubscript{\scriptsize $\pm$0.013} & 97.2\textsubscript{\scriptsize $\pm$0.017} 
& 71.3\textsubscript{\scriptsize $\pm$0.030} & \underline{69.3\textsubscript{\scriptsize $\pm$0.023}} & \textbf{60.5\textsubscript{\scriptsize $\pm$0.018}} & 55.6\textsubscript{\scriptsize $\pm$0.026} & \underline{66.4\textsubscript{\scriptsize $\pm$0.036}} \\
LLaMA4
& 66.9\textsubscript{\scriptsize $\pm$0.028} & 66.7\textsubscript{\scriptsize $\pm$0.028} & 69.2\textsubscript{\scriptsize $\pm$0.027} & 53.8\textsubscript{\scriptsize $\pm$0.031} & 96.7\textsubscript{\scriptsize $\pm$0.019}
& \underline{72.1\textsubscript{\scriptsize $\pm$0.020}} & 63.7\textsubscript{\scriptsize $\pm$0.014} & 46.5\textsubscript{\scriptsize $\pm$0.024} & \underline{62.2\textsubscript{\scriptsize $\pm$0.034}} & 37.1\textsubscript{\scriptsize $\pm$0.035} \\
GPT5 & 
68.4\textsubscript{\scriptsize $\pm$0.035} & 
68.0\textsubscript{\scriptsize $\pm$0.035} & 
71.1\textsubscript{\scriptsize $\pm$0.031} & 
55.9\textsubscript{\scriptsize $\pm$0.037} & 
97.8\textsubscript{\scriptsize $\pm$0.012} & 
70.4\textsubscript{\scriptsize $\pm$0.033} & 
62.4\textsubscript{\scriptsize $\pm$0.045} & 
44.9\textsubscript{\scriptsize $\pm$0.070} & 
57.2\textsubscript{\scriptsize $\pm$0.056} & 
37.3\textsubscript{\scriptsize $\pm$0.073} \\
Gemini2.5 & 
60.1\textsubscript{\scriptsize $\pm$0.028} & 
58.6\textsubscript{\scriptsize $\pm$0.031} & 
66.4\textsubscript{\scriptsize $\pm$0.026} & 
49.9\textsubscript{\scriptsize $\pm$0.027} & 
\textbf{98.9\textsubscript{\scriptsize $\pm$0.009}} & 
66.9\textsubscript{\scriptsize $\pm$0.034} & 
65.6\textsubscript{\scriptsize $\pm$0.033} & 
59.0\textsubscript{\scriptsize $\pm$0.033} & 
49.9\textsubscript{\scriptsize $\pm$0.022} & 
\textbf{72.6\textsubscript{\scriptsize $\pm$0.067}} \\
\midrule
\multicolumn{11}{l}{\textit{Training Free — MARS}} \\
Qwen2.5
& \underline{75.8\textsubscript{\scriptsize $\pm$0.021}}& \underline{75.8\textsubscript{\scriptsize $\pm$0.021}} & \textbf{75.8\textsubscript{\scriptsize $\pm$0.014}} & \underline{62.9\textsubscript{\scriptsize $\pm$0.024}} & 95.1\textsubscript{\scriptsize $\pm$0.013} & \textbf{75.9\textsubscript{\scriptsize $\pm$0.032}} & \textbf{71.3\textsubscript{\scriptsize $\pm$0.036}} & \underline{59.8\textsubscript{\scriptsize $\pm$0.049}} & \textbf{65.6\textsubscript{\scriptsize $\pm$0.029}} & 54.9\textsubscript{\scriptsize $\pm$0.076} \\
LLaMA4 
& \textbf{78.4\textsubscript{\scriptsize $\pm$0.033}} & \textbf{77.8\textsubscript{\scriptsize $\pm$0.033}} & 74.4\textsubscript{\scriptsize $\pm$0.034} & \textbf{68.6\textsubscript{\scriptsize $\pm$0.036}} & 81.4\textsubscript{\scriptsize $\pm$0.033}
& 70.3\textsubscript{\scriptsize $\pm$0.025} & 54.9\textsubscript{\scriptsize $\pm$0.038} & 28.6\textsubscript{\scriptsize $\pm$0.064} & 60.6\textsubscript{\scriptsize $\pm$0.098} & 18.9\textsubscript{\scriptsize $\pm$0.049} \\
GPT5 & 
75.1\textsubscript{\scriptsize $\pm$0.015} & 
75.0\textsubscript{\scriptsize $\pm$0.015} & 
\underline{75.5\textsubscript{\scriptsize $\pm$0.017}} & 
61.9\textsubscript{\scriptsize $\pm$0.024} & 
96.9\textsubscript{\scriptsize $\pm$0.017} & 
70.2\textsubscript{\scriptsize $\pm$0.016} & 
58.7\textsubscript{\scriptsize $\pm$0.022} & 
36.8\textsubscript{\scriptsize $\pm$0.039} & 
60.0\textsubscript{\scriptsize $\pm$0.044} & 
26.7\textsubscript{\scriptsize $\pm$0.036} \\
Gemini2.5 & 
72.3\textsubscript{\scriptsize $\pm$0.027} & 
72.1\textsubscript{\scriptsize $\pm$0.027} & 
73.5\textsubscript{\scriptsize $\pm$0.026} & 
59.3\textsubscript{\scriptsize $\pm$0.031} & 
96.8\textsubscript{\scriptsize $\pm$0.016} & 
70.5\textsubscript{\scriptsize $\pm$0.032} & 
63.7\textsubscript{\scriptsize $\pm$0.036} & 
48.2\textsubscript{\scriptsize $\pm$0.055} & 
57.4\textsubscript{\scriptsize $\pm$0.036} & 
41.9\textsubscript{\scriptsize $\pm$0.072} \\
\bottomrule
\end{tabular}
\end{table*}

\section{Methodology}
\subsection{Problem Formulation}


Given a video $V$ containing multimodal information including visual frames $F$ and audio transcription $\mathbf{A}$, where $F$ denotes the set of video frames, we uniformly sample a subset of frames $\mathbf{I} = \{I_1, I_2, \ldots, I_N\} \subseteq F$, where $N$ is a hyperparameter specifying the number of sampled frames. The task is to determine whether a video $V$ contains hateful content by prompting a VLM to output a label $y \in \{0,1\}$, where $y=1$ denotes hateful and $y=0$ non-hateful. The definitions of hateful content follow those adopted in the HateMM \cite{das2023hatemm} and MHC \cite{wang2024multihateclip} benchmarks.

Traditional training-based approaches learn from labeled datasets $\mathcal{D} = \{(V_i, y_i)\}_{i=1}^M$ by solving the optimization problem:
\[
\small
    f^* = \arg \min_{f \in \mathcal{F}} \mathcal{L}(f(V), y),
\]
where $\mathcal{F}$ represents the hypothesis space and $\mathcal{L}$ is the loss function. The operator $\arg \min$ specifies that $f^*$ is the function within $\mathcal{F}$ that minimizes the loss. In contrast, our framework operates in a training-free paradigm, leveraging VLMs for inference-time reasoning without parameter updates.

\subsection{Multi-Stage Adversarial ReaSoning}

Our framework adopts a four-stage process, explicitly evaluating competing hypotheses before reaching a final decision.

\subsubsection{Stage 1: Objective Content Representation}

The first stage establishes a neutral baseline through systematic content description. The stage prompt instructs the model to provide only a factual description without any interpretation (see Fig.~\ref{fig:framework} for prompts). We provide the model with a set of sampled video frames 
$\mathbf{I}$ together with the transcription $\mathbf{A}$. The model then produces an objective description:
\[
    D^{obj} = \Phi_{LLM}^{(1)}(\mathbf{I}, \mathbf{A} \mid P^{obj}),
\]
where $\Phi_{LLM}^{(1)}$ denotes the LLM reasoning function guided by the objective description prompt $P^{obj}$. 
The output $D^{obj}$ is a free-text factual description generated by the model. 

\subsubsection{Stage 2: Hate Content Hypothesis Inference}

Under the assumption: “the content contains hate speech,” the framework, guided by a prompt, $P^{hate}$, which explicitly requests hateful interpretation, identifies hate-supporting evidence $E^{hate}$, reasoning $R^{hate}$, and confidence $conf^{hate}$ (see Fig.~\ref{fig:framework} for prompt): 
\[
\begin{aligned}
    D^{hate} = \{E^{hate}, R^{hate}, conf^{hate}\} 
    &= \Phi_{LLM}^{(2)}\big(\mathbf{I}, \mathbf{A} \mid P^{hate}),
\end{aligned}
\]
where $P^{hate}$ encodes definitions and criteria for detecting hateful or offensive material, including group-based discriminatory language, dehumanizing symbols, and identity-based attacks. 
Here, $E^{hate}$ is returned as free-text evidence, $R^{hate}$ as free-text reasoning, and $conf^{hate}$ as a scalar confidence score $\in [0,1]$ provided by the model.

\subsubsection{Stage 3: Non-Hate Content Hypothesis Inference}

Conversely, under the assumption: “the content does not constitute hate speech,” the framework, guided by a prompt, $P^{non}$, which explicitly requests a non-hateful interpretation, collects non-hate evidence $E^{non}$, reasoning $R^{non}$, and confidence $conf^{non}$ (see Fig.~\ref{fig:framework} for prompt):
\[
\begin{aligned}
    D^{non}=\{E^{non}, R^{non}, conf^{non}\} 
    &= \Phi_{LLM}^{(3)}\big(\mathbf{I}, \mathbf{A} \mid P^{non}\big),
\end{aligned}
\]
This stage emphasizes distinguishing personal disputes from group targeting, considering contextual factors such as satire, artistic usage, or absence of discriminatory language patterns. 
Here, $E^{non}$ and $R^{non}$ are free-text outputs, while $conf^{non}$ is a scalar confidence $\in [0,1]$.

\subsubsection{Stage 4: Meta-Analytical Synthesis}

The final stage integrates competing hypotheses via structured meta-analysis. A synthesis function $\Psi$, guided by the meta-prompt $P^{meta}$ (see Fig.~\ref{fig:framework} for prompt), 
weighs the alternative hypotheses by evaluating evidence strength, contextual relevance, and potential harm. The outcome is expressed as a structured decision object:
\[
\begin{aligned}
D^{meta} &= \{\, (y_{pred},\; conf^{final}),\; K^{factors},\; R^{final} \,\} \\
         &= \Psi\big(\mathbf{I}, \mathbf{A}, D^{obj}, D^{hate}, D^{non} \mid P^{meta}\big).
\end{aligned}
\]
Here, $(y_{pred}, conf^{final})$ denotes the final decision label and its confidence score, where $y_{pred} \in \{0,1\}$ and $conf^{final} \in [0,1]$. $K^{factors}$ denotes the key factors determining the final judgement following comparative evidence analysis and $R^{final}$ signifies the rationale. The confidence score $conf^{final}$ is intended solely for interpretability rather than thresholding.
\begin{figure*}[t]
    \centering
    \includegraphics[width=0.8\textwidth]{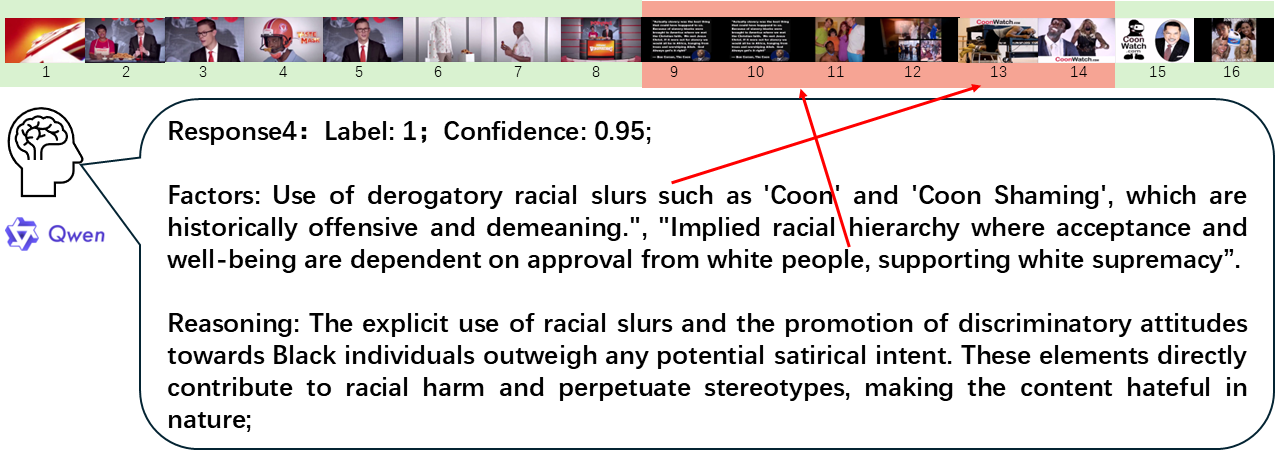}
    \caption{Examples of interpretability. 
    Red lines indicate corresponding factors and video content.
    }
    \label{fig:example}
\end{figure*}

\section{Experiments}
\label{sec:experiments}
\subsection{Experimental Setting}
We validate our framework using multiple VLM backbones: Qwen2.5-VL (32B) \cite{Qwen2.5-VL}, Llama4 (17B-128E) \cite{Llama4}, GPT5-mini \cite{GPT5}, and Gemini2.5-Flash \cite{Gemini2_5Flash}. Experiments are conducted with $N=16$, balancing hardware constraints and information coverage. Evaluation is performed on two hateful video benchmarks: HateMM (1,083 videos) and the Chinese subset of MHC (959 videos, slightly fewer than the 1,000 reported by MHC \cite{wang2024multihateclip} due to removals from Bilibili \cite{bilibili}). We adopt a 70/10/20 split for training, validation, and testing, with 5-fold cross-validation. Unlike prior work that relied on fixed splits with random seeds \cite{MoRE, wang2024multihateclip}, we ensure mutually exclusive test sets by repartitioning the dataset. For training-free inference, we evaluate the test set of each fold and report averaged results. In line with MHC \cite{wang2024multihateclip}, we merge hate and offensive labels into a single hate category for binary classification. All prompts were written in English for both datasets. Following standard protocols \cite{das2023hatemm, wang2024multihateclip}, we report Accuracy, Macro-F1, and Precision, Recall, and F1 for the hate class.

\subsection{Baseline Methods}
We compare our approach against several baseline methods:

Training-based: We incorporate training-based model results on these datasets, utilising the baseline model proposed by Das et al. \cite{das2023hatemm} and Wang et al. \cite{wang2024multihateclip}, alongside two state-of-the-art models: CMFusion\cite{CMFusion} and MoRE\cite{MoRE}. These models establish upper bounds for task-specific training performance and are strictly reproduced according to their original implementations and tested on new dataset splits.

Simple Prompting: This baseline prompts VLMs to classify videos, directly determining whether a video contains hateful content. The prompting method aligns with the zero-shot testing approach described in MHC \cite{wang2024multihateclip}.

CoT Prompting: 
We implement a CoT baseline requiring models to ``think step by step" about video content. Following established CoT frameworks by Xu et al. \cite{cot2024}, we first analysis video frames and text independently, then integrate multimodal information before outputting final judgments.

\subsection{Results and Discussion}
Table~\ref{tab:results} reports the performance comparison across two datasets. We note that MARS prioritizes precision over recall by design, reflecting real-world moderation requirements where false positives incur higher social and operational costs than false negatives. 

Our training-free \textit{MARS} framework consistently achieves higher precision, substantially reducing false positives and thereby improving overall accuracy. On the English dataset, it consistently outperforms the others baseline across all metrics and shows strong competitiveness against training-based methods. On the Chinese dataset, although all methods experienced performance degradation, \textit{MARS} maintained comparable accuracy while retaining a distinct precision advantage even at lower recall levels, achieving up to 7\% higher accuracy than training-based approaches. This suggests that \textit{MARS} is not only effective in low-resource multilingual settings but also resilient to domain shifts across languages. 

To demonstrate interpretability, we present a qualitative case study in Figure~\ref{fig:example}, showing the ability of \textit{MARS} to generate granular, evidence-based explanations. By elucidating the factors underlying hateful content classification, MARS enables rapid validation by content moderators and enhances system auditability, while mitigating hallucination risks through symmetric evidence construction and exposed intermediate reasoning.


We performed ablation experiments to assess each component of the framework as shown in Table~\ref{tab:component_ablation}. Removing either the objective description stage (w/o ObjDesc) or the assumption-based reasoning structure (w/o Assumption) reduced both accuracy and macro-F1, confirming the necessity of each stage in the reasoning pipeline. We further analysed robustness across varying model sizes and frame sampling strategies, as shown in Table~\ref{tab:parameters_ablation}. Frame sampling demonstrated stable performance, whilst model size exhibited significant improvements, indicating that \textit{MARS} can effectively scale with stronger backbones without sacrificing stability.

\begin{table}[t]
    \caption{Component ablation on Qwen2.5 (English dataset)}
    \centering
    \small
    \begin{tabular}{lcc}
        \toprule
        Configuration & Accuracy (\%) & Macro-F1 (\%) \\
        \midrule
        Full & 75.8 \textsubscript{\scriptsize $\pm$0.021} & 75.8 \textsubscript{\scriptsize $\pm$0.021} \\
        w/o ObjDesc & 72.9 \textsubscript{\scriptsize $\pm$0.021} & 72.8 \textsubscript{\scriptsize $\pm$0.021} \\
        w/o Assumption & 74.6 \textsubscript{\scriptsize $\pm$0.020} & 74.5 \textsubscript{\scriptsize $\pm$0.020} \\
        \bottomrule
    \end{tabular}
    \label{tab:component_ablation}
\end{table}

\begin{table}[t]
    \caption{Parameter analysis on Qwen2.5 (English dataset)}
    \centering
    \small
    \begin{tabular}{lcc}
        \toprule
        Configuration & Accuracy (\%) & Macro-F1 (\%) \\
        \midrule
        8 frames  & 74.4 \textsubscript{\scriptsize $\pm$0.026} & 74.3 \textsubscript{\scriptsize $\pm$0.027} \\
        16 frames & 75.8 \textsubscript{\scriptsize $\pm$0.021} & 75.8 \textsubscript{\scriptsize $\pm$0.021} \\
        32 frames & 76.2 \textsubscript{\scriptsize $\pm$0.017} & 76.1 \textsubscript{\scriptsize $\pm$0.017} \\
        \midrule
        7B  & 63.2 \textsubscript{\scriptsize $\pm$0.015} & 61.3 \textsubscript{\scriptsize $\pm$0.013} \\
        32B & 75.8 \textsubscript{\scriptsize $\pm$0.021} & 75.8 \textsubscript{\scriptsize $\pm$0.021} \\
        72B & 75.3 \textsubscript{\scriptsize $\pm$0.033} & 75.2 \textsubscript{\scriptsize $\pm$0.033} \\
        \bottomrule
    \end{tabular}
    \label{tab:parameters_ablation}
\end{table}


\section{Conclusion}
In this paper, we introduce \textit{MARS}, a training-free and interpretable framework for hateful video detection, that addresses three major limitations of prior approaches, namely reliance on scarce labeled data, lack of interpretability, and the high false-positive rates of VLMs. Experimental results demonstrate the superiority of \textit{MARS} over training-free baselines while maintaining competitive performance against training-based models. In addition, by incorporating explicit, evidence-based reasoning processes, \textit{MARS} offers greater prediction transparency and regulatory compliance while facilitating
human review, thereby laying the foundation for more auditable and trustworthy content moderation systems.


\bibliographystyle{IEEEbib}
\bibliography{strings,refs}

\end{document}